\documentclass{article}

\usepackage{arxiv}

\usepackage[utf8]{inputenc} 
\usepackage[T1]{fontenc}    
\usepackage{hyperref}       
\usepackage{url}            
\usepackage{booktabs}       
\usepackage{amsfonts}       
\usepackage{nicefrac}       
\usepackage{microtype}      
\usepackage{lipsum}
\usepackage{graphicx}
\graphicspath{ {./images/} }

\title{Text to Image Generation: Leaving no Language Behind }

\author{
 Pedro Reviriego \\
  Universidad Polit\'ecnica de Madrid\\
  28040 Madrid, Spain \\
  \texttt{pedro.reviriego@upm.es} \\
   \And
 Elena Merino-G\'omez \\
  Universidad de Valladolid\\
  47011 Valladolid, Spain \\
  \texttt{elena.merino.gomez@uva.es} \\
  \And
}

\begin{document}
\maketitle
\begin{abstract}
One of the latest applications of Artificial Intelligence (AI) is to generate images from natural language descriptions. These generators are now becoming available and achieve impressive results that have been used for example in the front cover of magazines. As the input to the generators is in the form of a natural language text, a question that arises immediately is how these models behave when the input is written in different languages. In this paper we perform an initial exploration of how the performance of three popular text-to-image generators depends on the language. The results show that there is a significant performance degradation when using languages other than English, especially for languages that are not widely used. This observation leads us to discuss different alternatives on how text-to-image generators can be improved so that performance is consistent across different languages. This is fundamental to ensure that this new technology can be used by non-native English speakers and to preserve linguistic diversity. 
\end{abstract}


\section{Introduction}

In the last decade, Artificial Intelligence (AI) has made significant breakthroughs and today outperforms humans in a number of tasks. As large datasets for training become available for many applications, and more sophisticated models are developed, AI is expected to overtake humans in more tasks in the coming years [1]. One application that has experience a dramatic improvement in the last years are text-to-image AI generators. These tools take as input a natural language description of an image in the form of a text prompt and generate an image that corresponds to what is described in the text. A good example are the DALL-E and Glide tools developed by OpenAI [2],[3],[4] or Imagen [5] and Parti [6] from Google. There are many other tools such as Cogview [7] and some are available as open-source projects like Craiyon (formerly DALL-E Mini) [8] or offer publicly available interfaces to generate images like MidJourney\footnote{Available at \url{www.midjourney.com}}  or DALL-E2 [3]. These text-to-image generators are attracting a lot of interest in many different communities ranging from computer science and graphic designers to artists, and even the general public. A good example of this interest is the use of DALL-E2 to create the front cover of a popular magazine\footnote{Available at \url{www.cosmopolitan.com/lifestyle/a40314356/dall-e-2-artificial-intelligence-cover/}}. This interest is expected to drive the development of this technology with new tools or new versions of existing tools being released in the coming months. These state of the art text-to-image generators are trained with huge datasets and have billions of parameters. For example, the latest Google generator, Parti can have up to 20 billion parameters and is trained on several billions of text/image pairs [6]. This means that the training of the generator requires a large computational cost and time.

Another area in which AI has also improved significantly is machine translation of texts [9]. Machine translation is now being used in many applications and is expected to be adopted in many others in the coming years [10],[11].  Machine translation is indeed a key technology to ensure that the latest technologies are available to humans regardless of their language thus ensuring diversity and inclusiveness [12]. Many tools have been implemented and are available for end-users or developers. There are indeed large efforts to make natural language processing in general and machine translation in particular, available in many languages. For example, BLOOM\footnote{Available at \url{https://huggingface.co/blog/bloom}} an advanced large language model with close to 200 billion parameters and that can generate text in 46 languages has been recently released as an open source project. As for text-to-image generators, training requires a huge computational effort, indeed more than 100 days running on a supercomputer were needed to train BLOOM. Another recent development is the No Language Left Behind (NLLB) initiative by Meta that provides translation among 200 different languages and for which the code has been released as open source [9].

As text-to-image generator technology emerges and consolidates, supporting language diversity and inclusiveness will become a priority. In this paper we take an initial step in this direction and analyze how current generators performance depends on the language used to input the text. We consider several generators and languages using simple texts to get a preliminary understanding on the state of the technology. These initial results show that for widely used languages, like Spanish, some of the generators have a similar performance to English, but others suffer a significant performance degradation. Instead, for languages used by smaller communities or no longer used, all the generators have a large performance degradation in most cases. Based on these results, we briefly discuss how language support can be achieved in text-to-image generators exploring different options. All of them seem to lead to an interaction between text-to-image generators and machine translation, opening an interesting topic for further research. 

The rest of the paper is organized as follows. In section 2 we describe how we have designed our initial evaluation experiments and report and discuss the results. Then in section 3 we briefly discuss different options to make the performance of text-to-image generators consistent across many languages. The paper ends with the conclusion in section 4.

\section{Initial Evaluation of Language Support in Text to Image Generators }
\label{sec:evaluation}

In order to have a preliminary idea of how different languages are supported on text-to-image generators, an initial evaluation has been done using four languages: 

\begin{enumerate}
    \item English. 
    \item Spanish. 
    \item Basque. 
    \item Latin.
\end{enumerate}

English is used as the benchmark for comparison; Spanish is included as a broadly used language that has also similarities with other Roman languages such as Italian, French or Portuguese; Basque\footnote{A "language isolate" used in parts of northern Spain that has no demonstrated links with any major language.} as an example of a language used by a small community that has no similarity with any major language; and Latin as an example of a language that is no longer used but has strongly influenced many widely used languages. The translation of the prompts has been done using Google translate and then checked and corrected when needed by native speakers and an expert in Latin.

In terms of generators, three commonly used and publicly available tools have been used:

\begin{enumerate}
    \item DALL-E2.
    \item MidJourney.
    \item Craiyon. 
\end{enumerate}

DALL-E2 is a state of the art text-to-image generator developed by OpenAI. DALL-E2 has been trained using several hundreds of millions of text/image pairs and uses an AI model with billions of parameters. The algorithms, datasets and performance are described in some detail in [3].  MidJourney does not disclose details about the implementation, but it is also publicly available and has issued several releases. In our test we used the implementation available during August 2022. Finally, Craiyon (formerly known as DALL-E Mini) is an open-source project that aims to reproduce the results of OpenAI's DALL-E with a simpler design. In more detail, Craiyon uses only 0.4 billion parameters in the model compared to the several billion of DALL-E2 which makes the training process much faster, and also reduces the storage requirements and the time needed to generate images. Craiyon was trained with three image/text datasets using in total approximately 15 million image/text pairs.

Finally, we have selected five random text prompts\footnote{The prompts with the translations used for each language are listed in an appendix at the end of the paper.} from the MS-COCO dataset [13] that have been used to illustrate the performance of DALL-E2 (see Figure 12 in [3]). These prompts are short and simple sentences and contain only common words and thus one would expect a text-to-image generator to be able to create images that relate to the text. This gives us a total of 60 different sets of images that are sufficient to illustrate how performance varies significantly across languages in current text-to-image generators. 

The results obtained for MidJourney are summarized in Figure 1. In this case, four images are shown per text prompt. It can be observed that for English, the images capture the meaning of the text input reasonably well. For Spanish part of the information in the text is lost and does not appear in the images. Finally, for Basque and Latin, the tool is unable to interpret the text and produces images that are completely unrelated to the meaning of the text. Therefore, it seems that this tool has a performance that depends strongly on the input language even for widely used languages.

The results for Craiyon are presented in Figure 2. This tool produces nine images per run with the default settings, the best image obtained is shown in the figure. In this case, the images for Spanish text prompts have a similar quality as those of English. Instead, results for Basque and Latin are bad, with Craiyon identifying part of the elements described in the text in the best cases and producing completely unrelated images in the rest. Finally, the results for DALL-E2 are shown in Figure 3. The tool generates four images by default and the best one is shown. It can be observed that the quality of the images is similar for English and Spanish. Instead, there is a significant performance loss for both Basque and Latin.

Therefore, this simple experiment confirms that current text-to-image generation technology does not provide consistent performance across languages even when using very simple text prompts. Indeed, some of the tools suffer a significant performance degradation even when using widely used languages such as Spanish. The degradation is dramatic for Basque and Latin. In the first case, probably few image/text pairs have been used to train the tools as it is not a widely used language. The second shows that current AI generators cannot infer the meaning of Latin texts even when they have been trained with multiple languages derived from it such as Spanish, Italian or French.

\begin{figure*}
    \centering
      \includegraphics[scale=1]{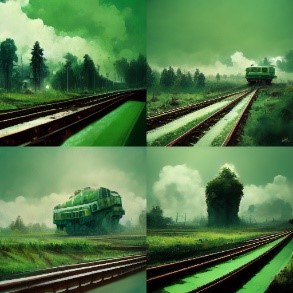}
      \includegraphics[scale=1]{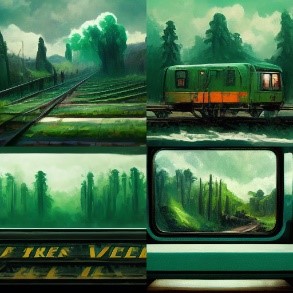}
      \includegraphics[scale=1]{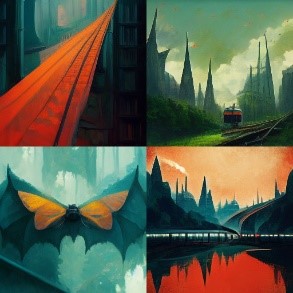}
      \includegraphics[scale=1]{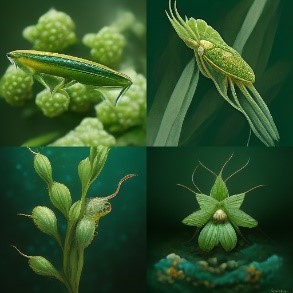}\\
      Text prompt 1: “a green train is coming down the tracks” (English, Spanish, Basque, Latin)\\
      \includegraphics[scale=1]{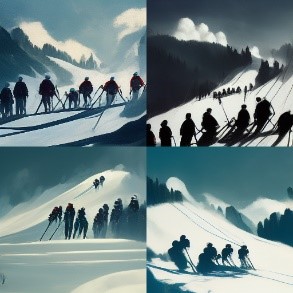}
      \includegraphics[scale=1]{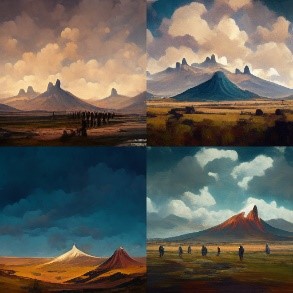}
      \includegraphics[scale=1]{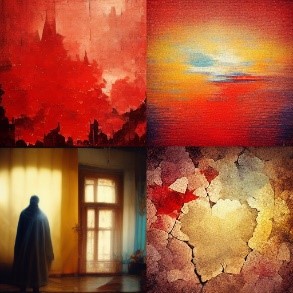}
      \includegraphics[scale=1]{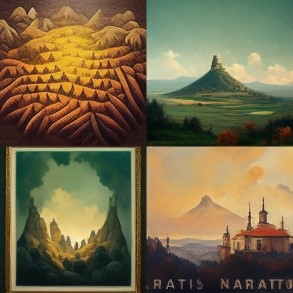}\\
      Text prompt 2: “a group of skiers are preparing to ski down a mountain” (English, Spanish, Basque, Latin)\\
      \includegraphics[scale=1]{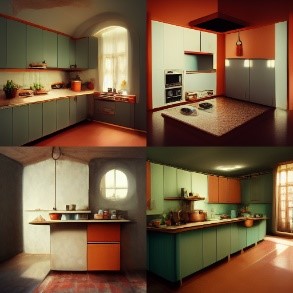}
      \includegraphics[scale=1]{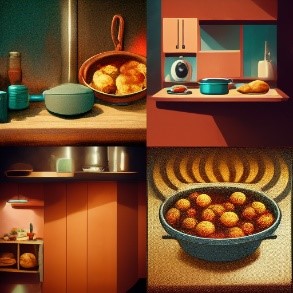}
      \includegraphics[scale=1]{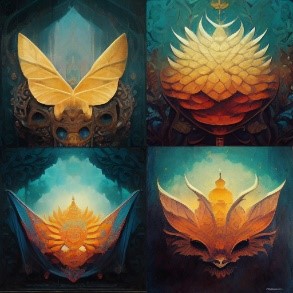}
      \includegraphics[scale=1]{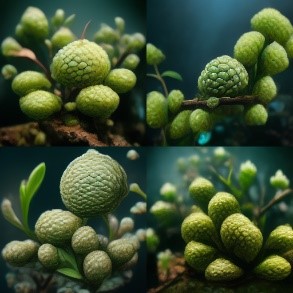}\\
      Text prompt 3: “a small kitchen with a low ceiling” (English, Spanish, Basque, Latin)\\
      \includegraphics[scale=1]{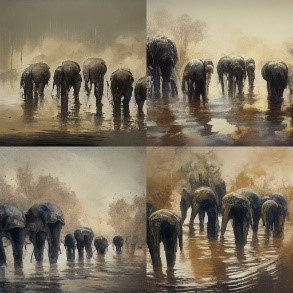}
      \includegraphics[scale=1]{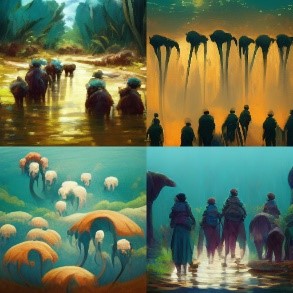}
      \includegraphics[scale=1]{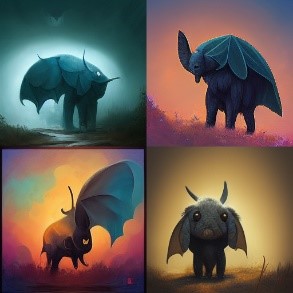}
      \includegraphics[scale=1]{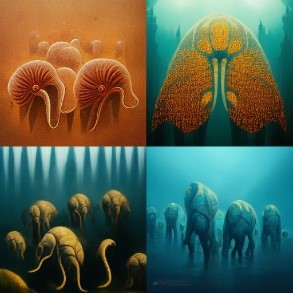}\\
      Text prompt 4: “a group of elephants walking in muddy water” (English, Spanish, Basque, Latin)\\
      \includegraphics[scale=1]{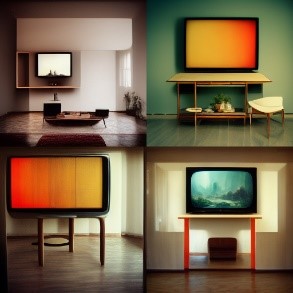}
      \includegraphics[scale=1]{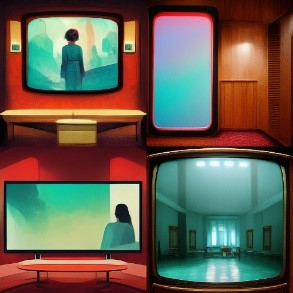}
      \includegraphics[scale=1]{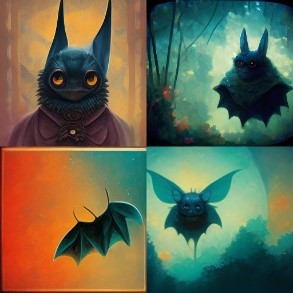}
      \includegraphics[scale=1]{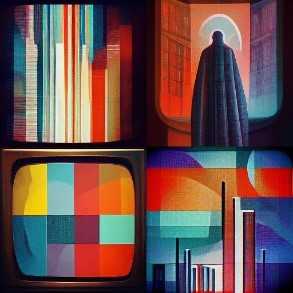}\\
      Text prompt 5: “a living area with a television and a table” (English, Spanish, Basque, Latin)\\
      
    \caption{\textbf{Results for MidJourney}}
\end{figure*}

\begin{figure*}
    \centering
      \includegraphics[scale=1]{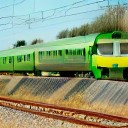}
      \includegraphics[scale=1]{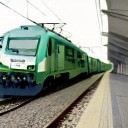}
      \includegraphics[scale=1]{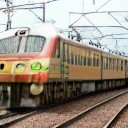}
      \includegraphics[scale=1]{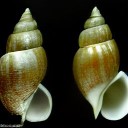}\\
      Text prompt 1: “a green train is coming down the tracks” (English, Spanish, Basque, Latin)\\
      \includegraphics[scale=1]{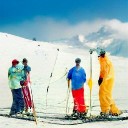}
      \includegraphics[scale=1]{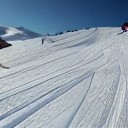}
      \includegraphics[scale=1]{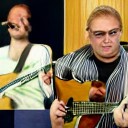}
      \includegraphics[scale=1]{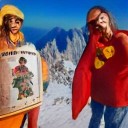}\\
      Text prompt 2: “a group of skiers are preparing to ski down a mountain” (English, Spanish, Basque, Latin)\\
      \includegraphics[scale=1]{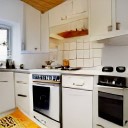}
      \includegraphics[scale=1]{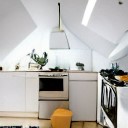}
      \includegraphics[scale=1]{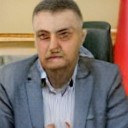}
      \includegraphics[scale=1]{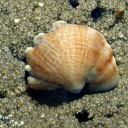}\\
      Text prompt 3: “a small kitchen with a low ceiling” (English, Spanish, Basque, Latin)\\
      \includegraphics[scale=1]{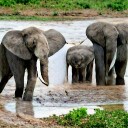}
      \includegraphics[scale=1]{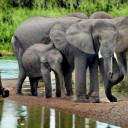}
      \includegraphics[scale=1]{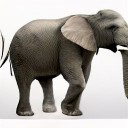}
      \includegraphics[scale=1]{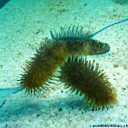}\\
      Text prompt 4: “a group of elephants walking in muddy water” (English, Spanish, Basque, Latin)\\
      \includegraphics[scale=1]{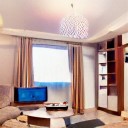}
      \includegraphics[scale=1]{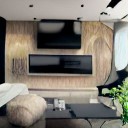}
      \includegraphics[scale=1]{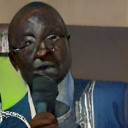}
      \includegraphics[scale=1]{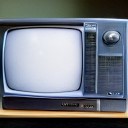}\\
      Text prompt 5: “a living area with a television and a table” (English, Spanish, Basque, Latin)\\
      
    \caption{\textbf{Results for Craiyon}}
\end{figure*}

\begin{figure*}
    \centering
      \includegraphics[scale=1]{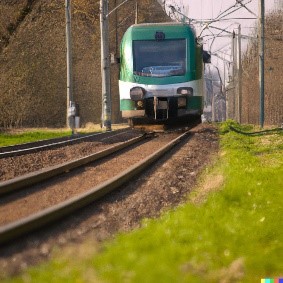}
      \includegraphics[scale=1]{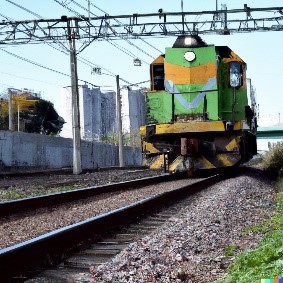}
      \includegraphics[scale=1]{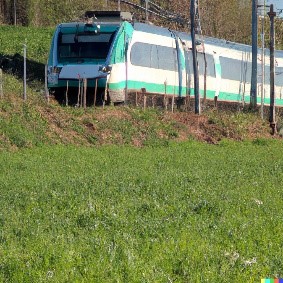}
      \includegraphics[scale=1]{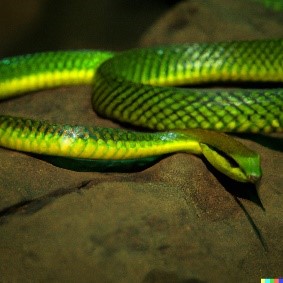}\\
      Text prompt 1: “a green train is coming down the tracks” (English, Spanish, Basque, Latin)\\
      \includegraphics[scale=1]{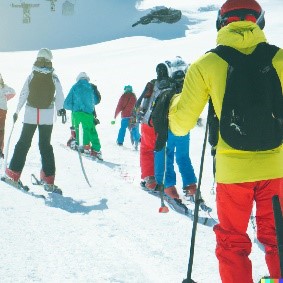}
      \includegraphics[scale=1]{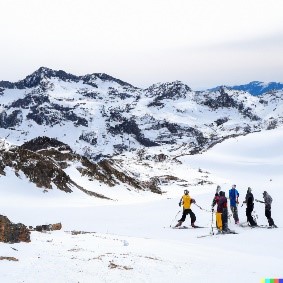}
      \includegraphics[scale=1]{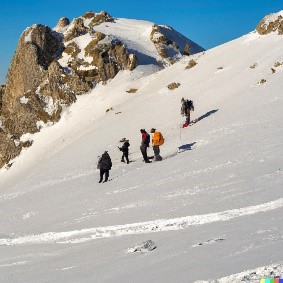}
      \includegraphics[scale=1]{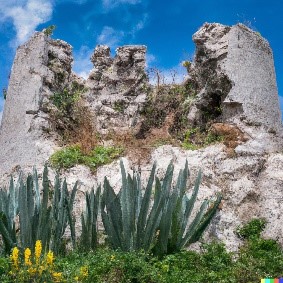}\\
      Text prompt 2: “a group of skiers are preparing to ski down a mountain” (English, Spanish, Basque, Latin)\\
      \includegraphics[scale=1]{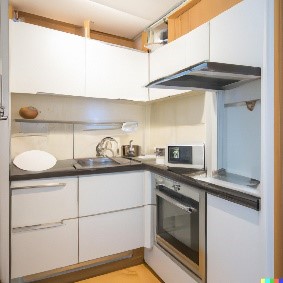}
      \includegraphics[scale=1]{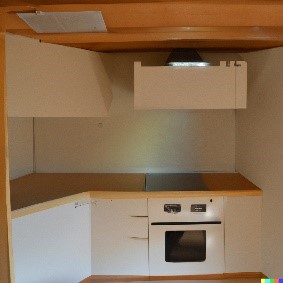}
      \includegraphics[scale=1]{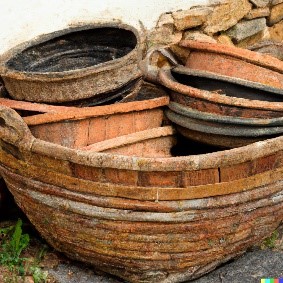}
      \includegraphics[scale=1]{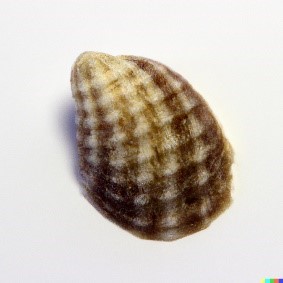}\\
      Text prompt 3: “a small kitchen with a low ceiling” (English, Spanish, Basque, Latin)\\
      \includegraphics[scale=1]{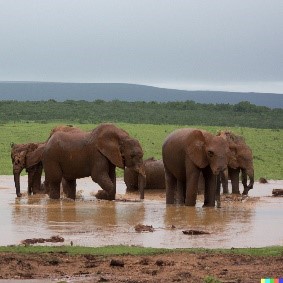}
      \includegraphics[scale=1]{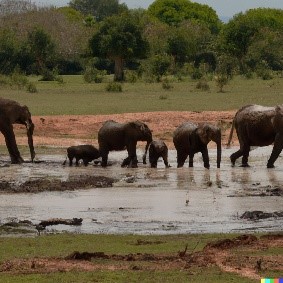}
      \includegraphics[scale=1]{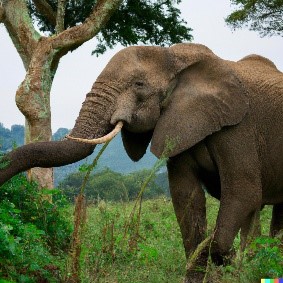}
      \includegraphics[scale=1]{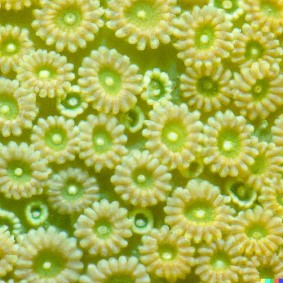}\\
      Text prompt 4: “a group of elephants walking in muddy water” (English, Spanish, Basque, Latin)\\
      \includegraphics[scale=1]{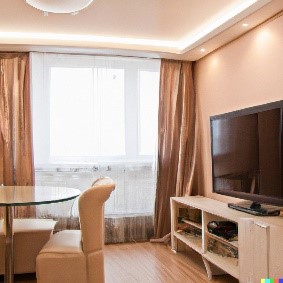}
      \includegraphics[scale=1]{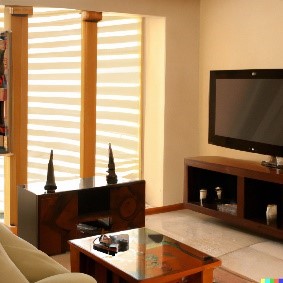}
      \includegraphics[scale=1]{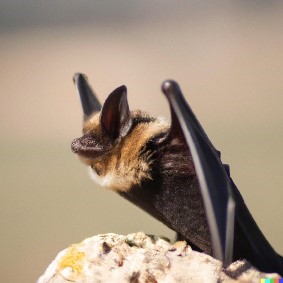}
      \includegraphics[scale=1]{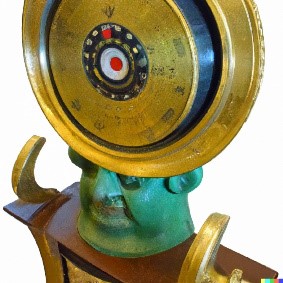}\\
      Text prompt 5: “a living area with a television and a table” (English, Spanish, Basque, Latin)\\
      
    \caption{\textbf{Results for DALL-E2}}
\end{figure*}

\section{Leaving no Language Behind in Text to Image Generation}
\label{sec:LNLB}

In this section we briefly discuss possible approaches to achieve similar performance across many languages in text-to-image generators. 

The first solution could be to include a relevant number of text/image pairs in all the target languages during the training of the generators. However, this has a number of issues: 1) getting those pairs does not seem to be straightforward and 2) training the model would be significantly more complex as the size of the dataset would grow by two orders of magnitude to cover a relevant number of languages. In more detail, there are some efforts to develop datasets of annotated images on several languages (see [14] and references within) but they are limited in the number of languages and number of images, very far from the hundreds of millions or billions used in state of the art text-to-image generators. A potential solution to the lack of annotated images for some languages could be to use automatic machine translation of the English annotations. Indeed, this has shown good performance in previous works such as [15]. However, even if we can generate many text/image pairs for all the target languages, we would still have to face the second issue, training the system with a much larger dataset.

A second approach would be to add a natural language processing module to the text-to-image generators that detects the language of the input text and when it corresponds to a language for which the system has poor performance applies machine translation to English and uses the generated text as input. This has a number of advantages compared to the previous solution: 1) the text-to-image generator can be trained on a much smaller dataset and 2) the machine translation is independent of the text-to-image generator so that advances in translation can be used as they become available or different implementations of the translation can be used with the same text-to-image generator. Therefore, it seems this second solution is more flexible and scalable. 

A final consideration is that the lack of human annotated images for many languages would in any case limit the ability of text-to-image generators to what can be achieved in translation (that has to be used either in the generation of the training dataset or to translate the input text) and thus minority languages would be somewhat lost in translation.

\section{Conclusion}
\label{sec:Conclusion}

In this paper we have performed an initial exploration of the performance of several recently developed text-to-image generators that are publicly available when using different languages for the text input. The results show that all generators suffer a significant degradation on the quality of their results when using languages that are not widely used and the degradation is even worse for classical languages such as Latin. Therefore, current generators do not seem to support language diversity and inclusiveness which are fundamental goals to make AI fair and responsible. Based on that observation, we have also discussed several alternatives to make the performance of text-to-image generators consistent across different languages. The analysis suggests that for low resources languages an attractive option is to use machine translation to preprocess the text and convert it to English. In any case, as text-to-image technology is still in its infancy, further evaluation and analysis is needed to fully understand how consistent performance across languages can be best achieved.

\newpage

\bibliographystyle{unsrt}  


\newpage

\textbf{APPENDIX: Text Prompts}

The text prompts used in the different languages (English, Spanish, Basque, Latin\footnote{For Latin, new words were taken from “Lexicon Recentis Latinitatis”. 2 vols. Vaticano: Libraria Editoria Vaticana, vol I: 1982; vol II: 1997 and Egger, Carolus. “Sermo Latinus Hodiernus”, Roma: Opus Fundatum “Latinitas”, 1986.}  ) are:

\begin{enumerate}
    \item "a green train is coming down the tracks”, "un tren verde viene por las vías", "tren berde bat dator trenbideetatik behera", "viridis hamaxósticus ferrivia venit"
    \item “a group of skiers are preparing to ski down a mountain”, "un grupo de esquiadores se prepara para esquiar montaña abajo", "eskiatzaile talde bat menditik behera eskiatzeko prestatzen ari da", "nartatorum turma parat nartis prolabi a monte" 
    \item “a small kitchen with a low ceiling”, “una pequeña cocina con un techo bajo”, "sukalde txiki bat sabai baxuarekin", "parva coquina humili tecto"
    \item “a group of elephants walking in muddy water”, "un grupo de elefantes caminando en agua fangosas", "elefante talde bat ur zingiratsuetan ibiltzen", "multitudo elephantorum ambulans in turbida aqua”
    \item “a living area with a television and a table”, "una sala de estar con un televisor y una mesa", "egongela bat telebista con mahai batekin" Ban eta "egongela bat telebista eta mahai batekin", "oecus televisione et mensa" 
    
\end{enumerate}

\end{document}